\documentclass[letterpaper]{article}

\usepackage{ijcai17}

\usepackage{times}

\usepackage{amsmath}

\usepackage{amssymb}
\usepackage{graphicx}
\usepackage{multirow}
\usepackage{bbm}
\usepackage{color}
\usepackage{amsfonts}
\usepackage{times}
\usepackage[small]{caption}

\title{Group-wise Deep Co-saliency Detection}

\author{Lina Wei$^1$, Shanshan Zhao$^1$, Omar El Farouk Bourahla$^1$, Xi Li$^{1,2,}$\thanks{corresponding author}, Fei Wu$^1$  \\
$^1$ Zhejiang University, Hangzhou, China \\
$^2$ Alibaba-Zhejiang University Joint Institute of Frontier Technologies, Hangzhou, China \\
\{linawzju, zsszju, obourahla, xilizju, wufei\}@zju.edu.cn
}

\begin{document}

\maketitle

\vspace{-5cm}
\begin{abstract}

  In this paper, we propose an end-to-end group-wise deep co-saliency detection approach to address the co-salient object discovery problem based on the fully convolutional network (FCN) with group input and group output. The proposed approach captures the group-wise interaction information for group images by learning a semantics-aware image representation based on a convolutional neural network, which adaptively learns the group-wise features for co-saliency detection. Furthermore, the proposed approach discovers the collaborative and interactive relationships between group-wise feature representation and single-image individual feature representation, and model this in a collaborative learning framework. Finally, we set up a unified end-to-end deep learning scheme to jointly optimize the process of group-wise feature representation learning and the collaborative learning, leading to more reliable and robust co-saliency detection results. Experimental results demonstrate the effectiveness of our approach in comparison with the state-of-the-art approaches.

\end{abstract}

\begin{figure*}[htb]
\vspace{0.2cm}
\begin{center}
\includegraphics[width=1\textwidth]{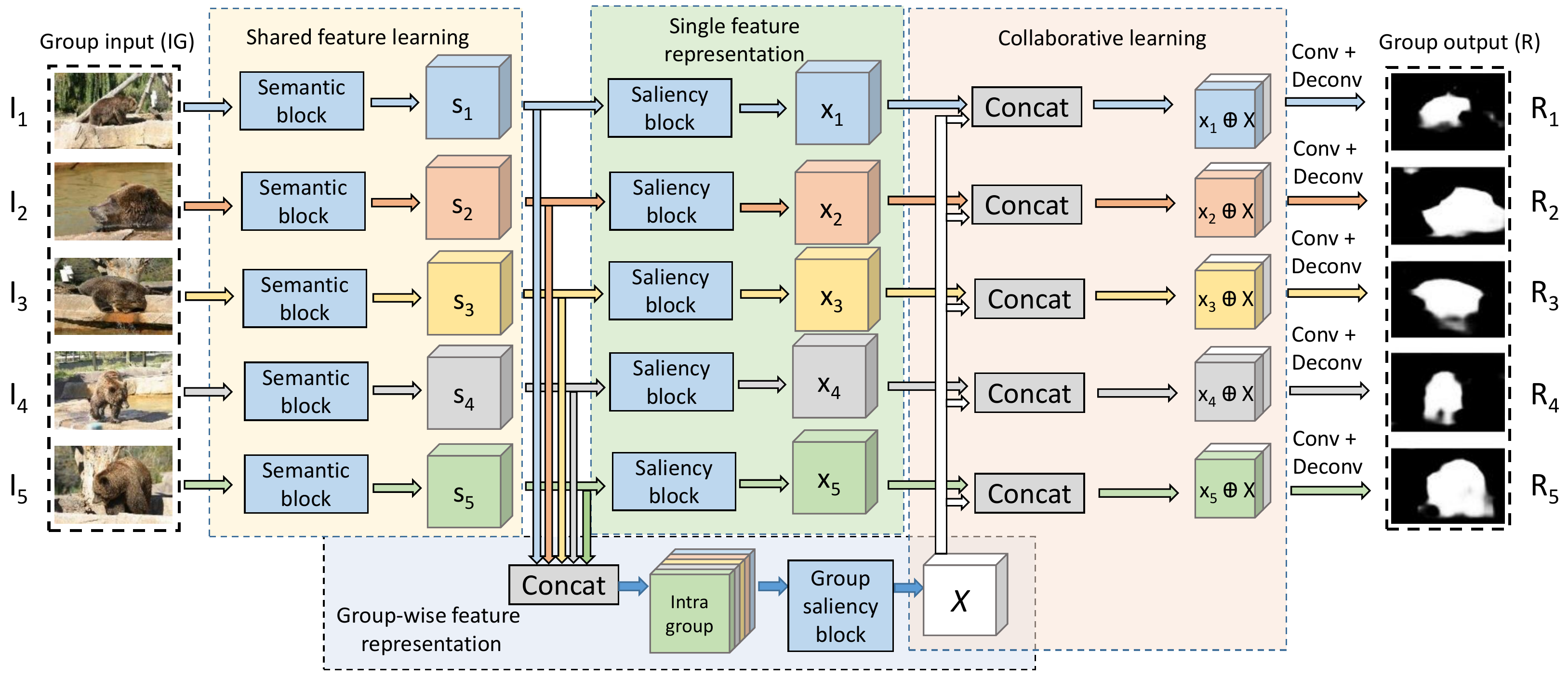}
\end{center}
\vspace{-1.5em}
\caption[]{Illustration of the proposed network architecture for co-saliency detection. The group images $\{I_{1},I_{2},...,I_{5}\}$ first go through the semantic feature extraction block, its results which are the image features $\{s_{1},s_{2},...,s_{5}\}$, as well as a concatenated version of them are then passed through the (group) saliency feature extraction block, then, the result $X$ coming from the concatenated version is further concatenated to the results coming from each of the other individual results$\{x_{1},x_{2},...,x_{5}\}$, then a convolution + deconvolution is applied on each to recover the saliency. $\bigoplus$ in this figure means concatenate operation}
\label{fignet}
\end{figure*}
\section{Introduction}

In principle, co-salient object detection~\cite{batra2010icoseg,chang2011co,li2011co,fu2013cluster,li2013co} is defined as the problem of discovering the common and salient foregrounds from an image group containing multiple images at the same time. It has a wide range of applications on computer vision tasks, such as image or video co-segmentation~\cite{fu2015object1,fu2015object,wang2015saliency}, object localization~\cite{tang2014co,cho2015unsupervised}, and weakly supervised learning~\cite{siva2013looking}.

In order to detect co-salient regions precisely, we need to focus on two key points: 1) how to extract effective features to represent the co-salient regions; 2) how to model the interactive relationship between images in a group to obtain the final co-saliency maps. For 1), feature representation in the co-saliency detection task should not only reflect the individual properties of each image itself, but also express the relevance and interaction between group images. For 2), we know that images within a group are contextually associated with each other in different ways such as common objects, similar categories, and related scenes. The co-saliency detection job tries to use this information to find the target saliency maps, so we can utilize the consistency information within these image groups and capture an interaction between the images so that they mutually reinforce and enhance each other's saliency regions.

For tackling lots of challenges, we need to design a model that can extract robust features that reflect the individual properties of each image as well as features that represent the group-wise information such as group consistency, object interactions and, to a minor extent, the objects that are present in only single images but not the rest of the images. A series of approaches have been proposed from different points of view. Some methods~\cite{chang2011co,fu2013cluster,li2013co,cheng2014salientshape} consider that the co-salient objects appearing in the group images should share a certain consistency in both low-level feature and high-level semantic feature~\cite{zhang2016detection,zhang2015self,zhang2016cosaliency}, however, they do not model the interaction between the group-wise features and single image features, which can contain information that can improve the results. Some approaches detect the single-image individual saliency and the common salient regions of a group in a separate manner~\cite{ge2016co,li2013co} and, they also detect the intra-image and inter-image saliency separately from other information priors, such as the objectness~\cite{li2014co,liu2014co}, the center priors~\cite{chen2014implicit}, and the border connectivity~\cite{ye2015co}.
Usually, calculating the intra-image and inter-image saliency separately is incapable of well capturing the intrinsic semantic
interaction information among images within each group, which is important to the co-saliency detection quality.

Motivated by this observation, we propose a co-saliency deep model based on a fully convolutional network(FCN) with group input and group output. Our aim is to make use of all the information available and create a robust and effective network. Our model needs to take into account both the image properties and the intra group information while processing the co-saliency results. We design our network to be fully convolutional, this allows it to fully benefit from the local relationships between the pixels in an image, it is also designed deep enough to have a large receptive field. The network will extract the semantic features of the images, then will be divided into two branches. Namely, one processes each image individually and the other takes into account all the image group, the branches are later merged. This allows the network to learn features not only from the individual image properties, but also from the intra group properties, leveraging the shared and unique information between the images, resulting in accurate co-saliency maps. Our deep model takes a data-driven learning pipeline for capturing the collaboration and consistency intra image group, and is trained end-to-end.

The main contributions of this work are summarized as follows:

First, we propose a unified group-wise deep co-saliency detection approach with group input and group output, which takes advantage of the interaction relationships between group images. The proposed approach performs feature representation for both single image(e.g., individual objects and unique properties) and group consistency (e.g., common background and similar foreground), which generally leads to an improvement in the performance of the co-saliency detection.

Second, we set up an end-to-end deep learning scheme(FCN) to jointly optimize the process of group-wise feature representation learning and the collaborative learning, leading to more reliable and robust co-saliency detection results. The collaborative learning process combines the group-wise saliency and single image saliency in a unified framework that model a better interaction relationships between group images.

\section{Proposed Approach}

\subsection{Problem Formulation}
Given a group of images $IG = \{I_{i}\}_{i=1}^{K}$ where $K$ is the number of images in group. Our goal is to discover the co-salient regions $R = \{R_{i}\}_{i=1}^{K}$ for this image group, where $R_{i}$ is the saliency region for image $I_{i}$. In a simple saliency problem, each saliency region depends on its image and so we theoretically wish to find $R_{i}$ such that :

\begin{equation}\label{eq:Rionly}
\begin{aligned}
R_{i} = f(I_{i}; \Theta)
\end{aligned}
\end{equation}
where $f$ is a regression function that takes the image $I_{i}$ as input, and outputs the desired saliency map by learning a set of parameters $\Theta$. However, in our case, $I_{i}$ exists within a group of images that are contextually associated with each other, so that each saliency region has a certain interaction and depends on that of the other images. This changes the function we want to find to :

\begin{equation}\label{eq:Ri}
\begin{aligned}
R = F(IG; \Theta)
\end{aligned}
\end{equation}

In order to formulate the framework, we propose an end-to-end FCN with group input and group output which will process all the images at the same time and combine them at the feature level covering the needed theoretical necessity of taking into account all the images. The proposed group-wise co-saliency detection approach mainly consists of two components: 1) encoding the group images into co-feature representations by group-wise and single image feature learning to better obtain the comprehensive information, 2) collaborative learning by combing the group-wise feature with the single image feature through a unified joint learning structure which can comprehensively preserve common objects of the group and unique information for the single image. The architecture of the proposed approach is shown in Figure~\ref{fignet}.

\begin{figure}
\begin{center}
\includegraphics[width=.95\linewidth]{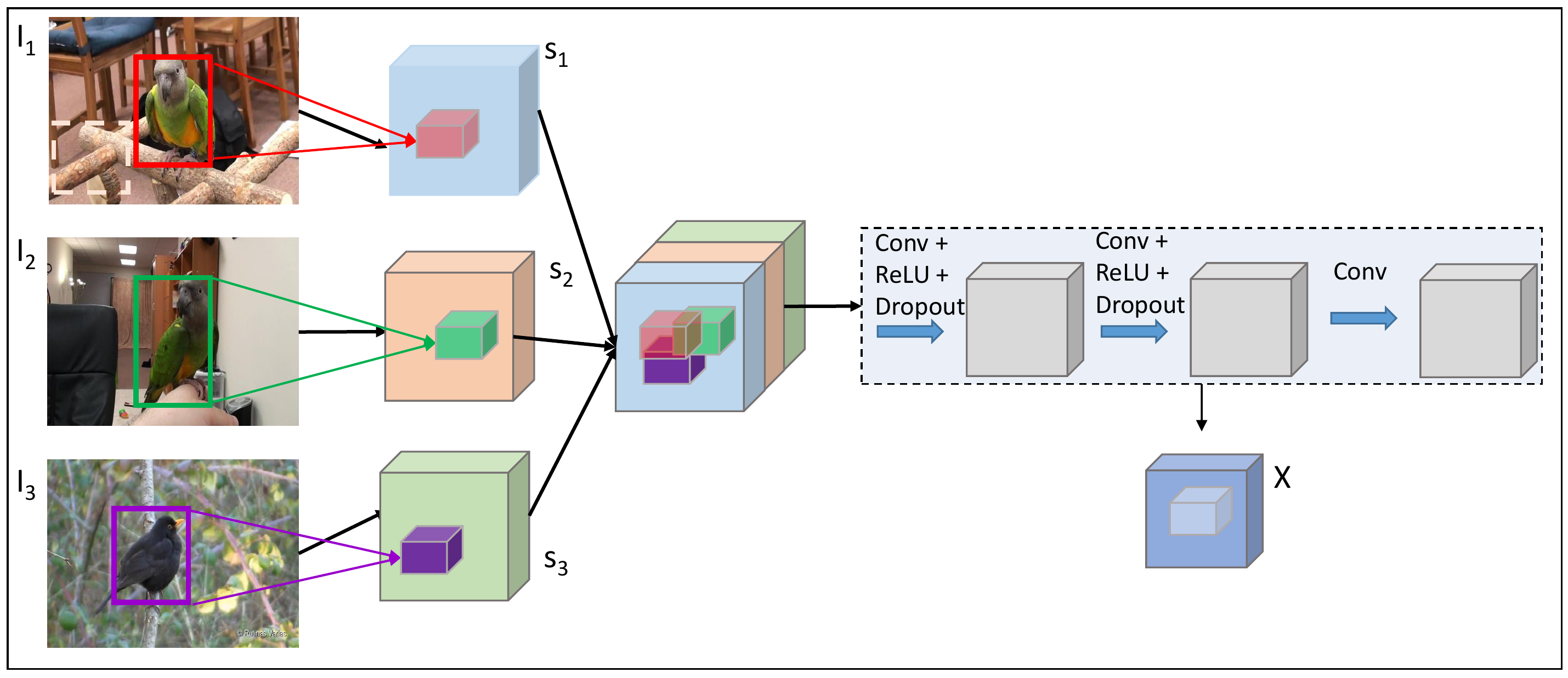}
\end{center}
\caption[]{Illustration of the working mechanism of group-wise feature learning by fusion strategy. The bounding boxes with different colors indicate the birds appear in different images and they are correlated with each other. We solve the problem by three stages: 1) group wise consistency investigation by group feature interaction; 2) joint aggregation by feature concatenation; and 3) saliency computation by a convolutional process.}
\label{figgm}
\end{figure}

\subsection{Semantic Image Representation}

In co-saliency detection, image representation is facing a number of challenges, such as multiple objects, occlusion, and diverse background. More importantly, co-feature representation for image group is the emphasis of our framework. It mainly consists of two component: first, constructing the group-wise feature representation which takes advantage of the intra group theoretical consistency to better obtain the interaction information of group images; and second, computing the single image semantic feature representation for each image individually.

\vspace{6pt}
\subsubsection{Group-wise Feature Representation}
As shown in Figure~\ref{fignet}, we adopt a group input and group output FCN to model the group semantic information for a joint representation. The initial high-level semantic feature $s_{i}$ for each image $I_{i}$ parameterised by $\Theta_{shared}$:

\begin{equation}\label{eq:si}
\begin{aligned}
s_{i} = f_{shared}(I_{i}; \Theta_{shared})
\end{aligned}
\end{equation}
where $f_{shared}$ is a convolutional process representing shown as the ``semantic block"  in Figure~\ref{fignet}, this block has $13$ convolutional layers, it has the parameters $\Theta_{shared}$ which are shared among all the  semantic blocks. With group input, we generate the shared feature $s = \{s_{i}\}_{i=1}^{K}$ for each image and these features will be the base on which we will do the next steps, and will be the link between the individual and intra group features since both use it.

Given the image group $IG$, the problem of group-wise feature representation is converted to the task of how to correspond the related components(such as common objects) defined in a group by their initial feature maps and to learn the interaction between images based on the group consistency.

The next step is concatenating these shared features and then applying $3$ convolution layers(shown in Figure~\ref{figgm}), this will give the network the possibility to extract the necessary group-wise information that can later be used for the computation of the saliency maps, it is defined as :

\begin{equation}\label{eq:XG}
\begin{aligned}
X &= f_{intra}(s; \Theta_{intra}) \\
\end{aligned}
\end{equation}
where $\Theta_{intra}$ is the parameters learned from $3$ convolutional layers and $f_{intra}$ is the convolutional process representing combining with the concatenation of $s = \{s_{i}\}_{i=1}^{K}$ shown in the ``group saliency block" of Figure~\ref{fignet}.

\vspace{6pt}
\subsubsection{Single Image Feature Representation}
The single image features $x = \{x_{i}\}_{i=1}^{K}$ encode the individual properties for each image $I_{i}$. As shown in the ``Single feature representation" of Figure~\ref{fignet}, taking advantage of the FCN, the feature is generated by a $3$ convolutional-layer network. It is defined as follows:

\begin{equation}\label{eq:x}
\begin{aligned}
x_{i} &= f_{single}(s_{i}; \Theta_{single}) \\
\end{aligned}
\end{equation}
where $\Theta_{single}$ are the parameters learned from the convolutional process $f_{single}$.

\begin{figure}
\begin{center}
\includegraphics[width=.95\linewidth]{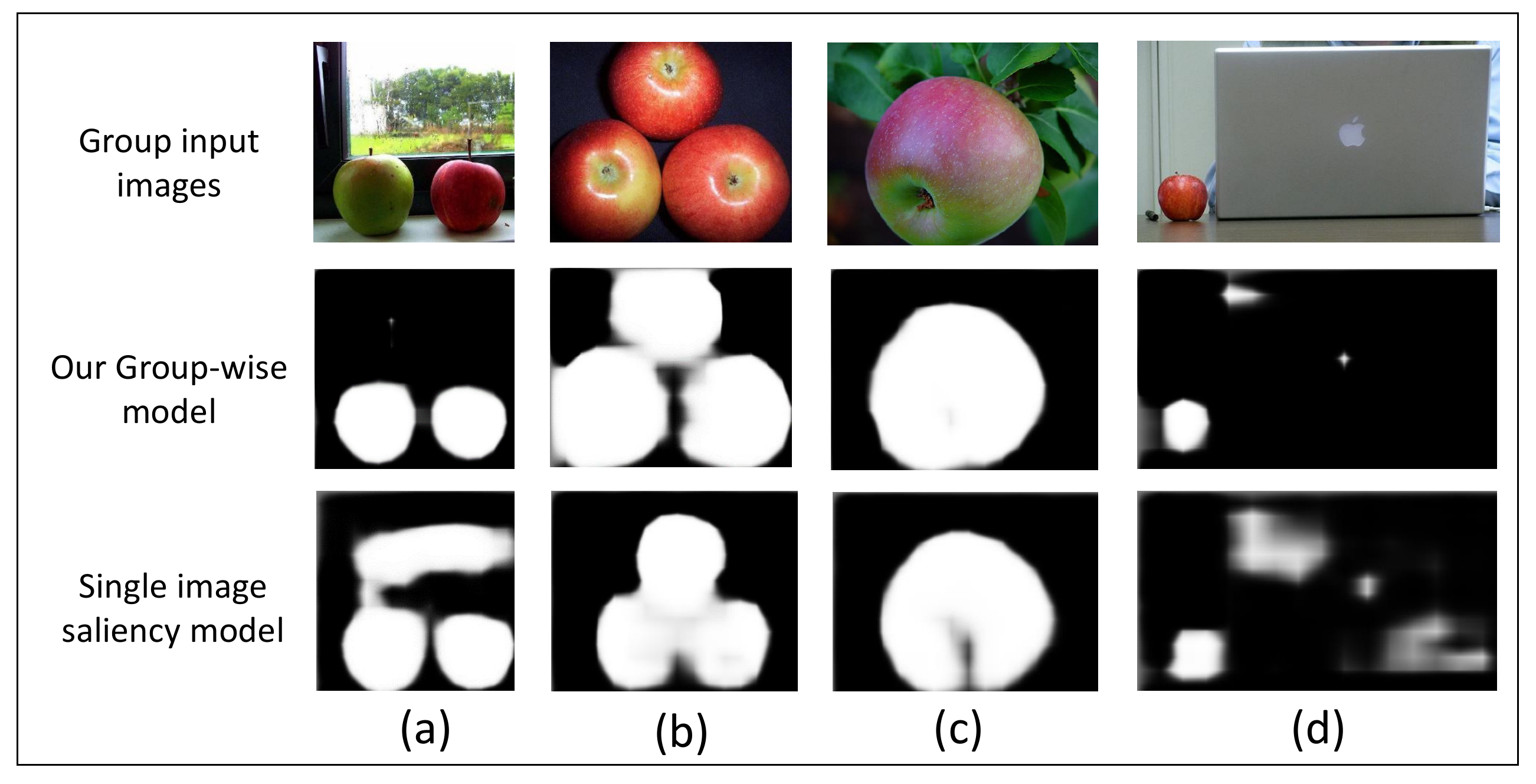}
\end{center}
\caption[]{Comparison of our group-wised model and the single image saliency model. With the group-wised feature representation, our model enhanced the common objects and weakened the other parts.}
\label{figgf}
\end{figure}

Applying these $3$ convolutional layers results in deeper features of each image. These are the features that will be combined with the group-wise features extracted in the previous sub-section. The merging is important because as shown in Figure~\ref{figgf}, some objects can be salient, but not present in the entire group, like the tree visible from the window in Figure~\ref{figgf} (a). This shows the necessity of the merging of the two features give the network the necessary flexibility so that it can weaken the saliency map in the regions where a salient object is not present in all the group. The other reason for the merging is enhancing the salient regions for objects that are present in all the image group, this is illustrated by Figure~\ref{figgf} (b) and (c) where the apples which are present in all the images have an increased saliency degree in a group-wise model than in a single image model.

\begin{figure*}
\vspace{0.2cm}
\begin{center}
\includegraphics[width=1.\linewidth]{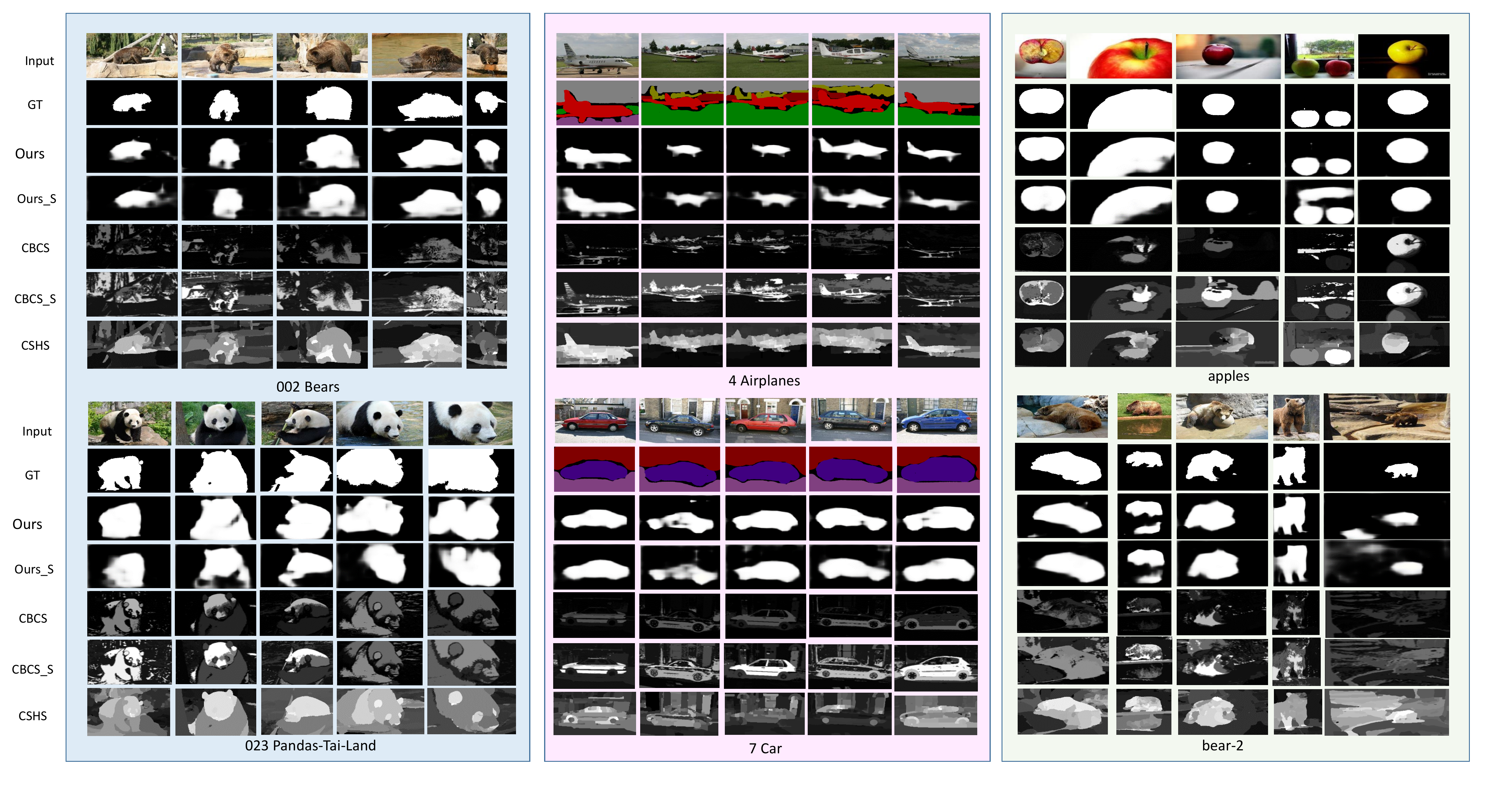}
\end{center}
\vspace{-1.5em}
\caption[]{Visual comparison of co-saliency detection on three benchmark datasets. From left to right, the examples are from iCoseg dataset, MSRC dataset, and Cosal2015 dataset, respectively. Obviously, the proposed method performs well in these datsets.}
\label{figsaliency}
\end{figure*}

\subsection{Collaborative Learning for Image Group}

As described previously, we construct the collaborative learning strategy from two components: the group-wise feature learning and the single image individual feature learning, which aims to adaptively capture the interaction relationships between group images and meanwhile retain the characteristics of single image itself. As shown in Figure~\ref{fignet}, the collaborative learning structure is discovered through joint learning for $x$ and $X$. Specifically, that means the common object regions are activated by convolutional process and the unique characteristics of single image are weakened but still retained for the final saliency estimation. The merging is defined as:

\begin{equation}\label{eq:R}
\begin{aligned}
R = f_{collaborative}(x,X; \Theta_{collaborative}) \\
\end{aligned}
\end{equation}
where $f_{collaborative}$ is the function that concatenates each $x_{i}$ with $X$, and then applies a convolutional and a deconvolutional layer on each of the results, which gives us the final group saliency, this is illustrated by the ``collaborative learning" part of Figure~\ref{fignet}, this architecture allows the network to combine the single image features and the group-wise features and obtain the saliency from their combined information.

\subsection{Training}

In principle, image representation and the learning strategy are correlated and complementary problems which can mutually promote each other. Thus we develop a unified end-to-end data-driven framework with group input and group output, where the group-wise feature and the single image features are learned jointly and adaptively in a supervised setting through the architecture illustrated in Figure~\ref{fignet}. For training, all the parameters $\Theta$ are learned via minimizing a loss function, which is computed as the errors between the saliency map and the ground truth. Let $\{IG_{i}\}_{i=1}^{N}$ and $\{GT_{i}\}_{i=1}^{N}$ denote a collection of training samples where $N$ is the number of image groups. Our network is trained by minimizing the following cost function:

\begin{equation}\label{eq:net}
\begin{aligned}
\sum_{i=1}^{N}
\|(GT_{i}-g(IG_{i}; \Theta)\|_F^2
\end{aligned}
\end{equation}
where $\Theta = \{\Theta_{shared},\Theta_{single},\Theta_{intra},\Theta_{collaborative}\}$, $g$ is the function that, given an input group, outputs the corresponding saliency maps for it. This cost function corresponds to the squared Euclidean loss term. The network is trained by the stochastic gradient descent (SGD) method to minimize the above cost function, a regularization is applied on all the training samples and all the parameters are learned simultaneously.

\section{Experimental Results}

\subsection{Experimental Setup}

\vspace{6pt}
\subsubsection{Datasets}

In order to evaluate the performance of the proposed approach, we conduct a set of qualitative and quantitative experiments on three benchmark datasets annotated with pixel-wised ground-truth labeling, including the iCoseg
dataset~\cite{batra2010icoseg}, the MSRC-v2 dataset~\cite{winn2005object} and the Cosal2015 dataset~\cite{zhang2016detection}. The iCoseg dataset contains $643$ images which divided into $38$ groups and they are challenging for co-saliency detection task because of the complex background and multiple co-salient objects. Note that we only use subset5 in this dataset which contains $5$ images in each group. Another large dataset widely used in co-saliency detection is the MSRC-v2 datasets which contains $591$ images in $23$ object classes with manually labeled pixel-wise ground truth data. It is more challenging than iCoseg dataset because of the diverse colors and shapes. The cosal dataset contains 50 image groups and totally 2015 images which are collected from challenging scenarios in the ILSVRC2014 detection benchmark~\cite{russakovsky2015imagenet} and the YouTube video set~\cite{prest2012learning}.

\vspace{6pt}
\subsubsection{Implementation Details}

The fully convolutional network (FCN) is implemented by using the Caffe~\cite{jia2014caffe} toolbox. We initialize our network by using a pretrained version of the single image input network (over the MS COCO dataset) which is based on the VGG 16-layer net~\cite{simonyan2014very}
and then, transfer the learned representations by fine-tuning~\cite{donahue2014decaf} to the co-saliency task by group input and group output. We construct the deconvolution layers by upsampling, whose parameters are initialized as simple bilinear interpolation parameters and iteratively updated during training. We resize all the images and ground-truth maps to $128\times 256$ pixels for training. The momentum parameter is chosen as 0.99, the learning rate is set to 1e-10, and the weight decay is 0.0005. We need about 60000 training iterations for convergence.

The training data we used in our approach are generated from existing image dataset(Coco dataset~\cite{lin2014microsoft}) which has $9213$ images with the masks information. In the proposed network, we set up the number of images in each group to $5$, namely, $K=5$. Following the approach of~\cite{siva2013looking}, we extract Gist and Lab color histogram features, and then calculate the Euclidean distance between images to find $4$ other images that are most similar to each one. In this way, we make up training groups. For testing, we randomly sample $5$ images from each group as the new image group to ensure the group input size to our model. This sampling procedure proceeds to generate a set of new
image groups (with the cardinality being 5) until all the original images are covered in the generated new image groups.
For iCoseg dataset, we directly adopt the subset~\cite{batra2010icoseg} which contains $5$ images in each group.
\begin{figure*}
\vspace{0.2cm}
\begin{center}
\includegraphics[width=1\textwidth]{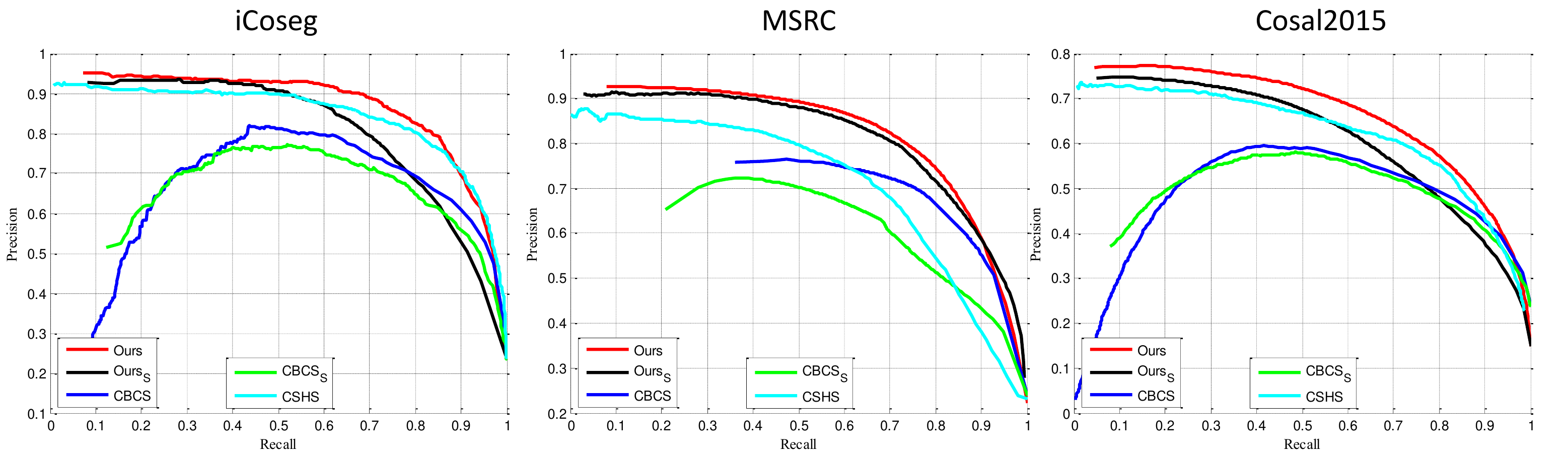}
\end{center}
\caption[]{Precision-recall curves of different saliency detection methods on 3 benchmark datasets. Overall, the proposed approach performs well with higher precision in the case of a fixed recall.}
\label{PR}
\end{figure*}

\subsection{Evaluation Metrics}

In the experiments, we utilize four metrics for quantitative performance evaluations, the Precision and Recall (PR) curve, F-measure, mean absolute error (MAE). Specifically, the PR curve reflects the object retrieval performance in precision and recall by binarizing the final saliency map using different thresholds (usually ranging from 0 to 255)\cite{borji2015salient}.
The F-measure characterizes the balance degree of object retrieval between precision and recall such that: $F_{\eta}=\frac{(1+\eta^{2}) Precision \times Recall}{\eta^{2}\times Precision + Recall}$ where $\eta^{2}$ is typically set to $0.3$ like the most existing literature work.
%
%
In addition, MAE refers to the average pixel-wise error between the saliency map and ground truth.
Finally, AUC evaluates the object detection performance, and computes the area under the standard ROC curve (false positive rate and true positive rate).

\subsection{State-of-the-Art Performance Comparison}

In the experiments, we compare the proposed approach with several representative
state-of-the-art methods including CSHS~\cite{liu2014co} and CBCS~\cite{fu2013cluster},
whose source codes are publicly available.
To investigate the performance differences with and without group interactions, we also make a comparison with
the saliency detection approaches for our work and CBCS without group interactions, respectively referred to as
Ours\_S and CBCS-S.
The experimental results are shown in Figure~\ref{figsaliency}. These examples belong to $6$ groups of the $3$ datasets mentioned above. From the comparison of these examples, we can observe that our proposed approach can better capture the common (in semantic-level) object regions, it also gives more clear borders between the salient and non-salient regions. As shown by the results on the iCoseg image groups which are illustrated on the left set (blue) of Figure~\ref{figsaliency}. The proposed approach does a better job on separating the salient regions and the background with clear boundaries. The middle set (pink) shows the groups in MSRC datast which is mainly for segmentation task. The co-saliency model captures the common objects well, in the semantic level. The right set (green) from Cosal2015 dataset is more challenging that the common objects in this dataset are always in shapes, colors, and viewed from different perspectives.
Therefore, our approach performs better than the competing approaches in most cases. Moreover,
the proposed group-wise approach with
group interactions gives rise to the performance gains relative to the corresponding approach without group interactions.

For quantitative comparison, the PR-curve is shown in Figure~\ref{PR} on the three datasets, it is observed that our approach performs best in all the datasets. Table~\ref{tab:comp} shows the comparison between the approaches through different evaluations. In iCoseg dataset and MSRC dataset, the proposed approach performs better than others on most evaluations.
In the challenging dataset Cosal2015 (with very complex scene clutters), the proposed approach performs best on all evaluations.

\begin{figure}
\vspace{0.2cm}
\begin{center}
\includegraphics[width=1\linewidth]{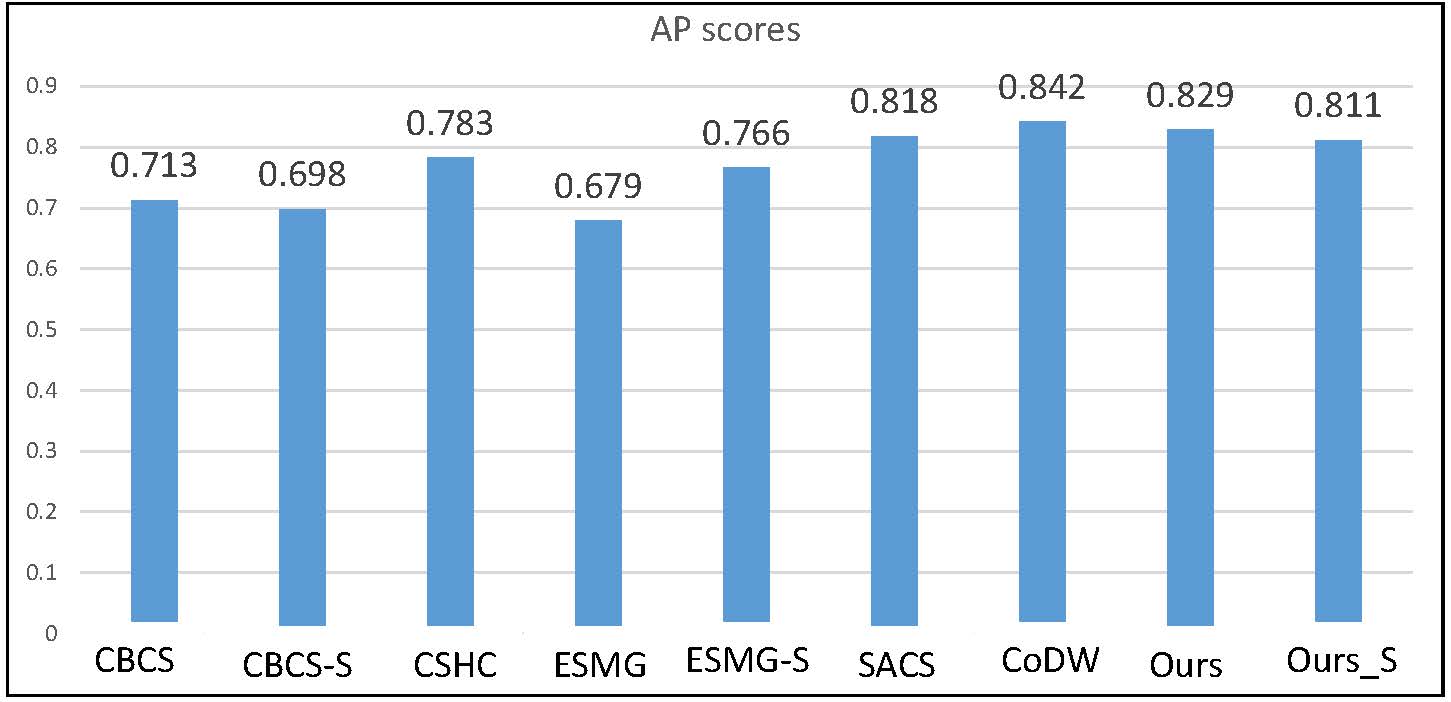}
\end{center}
\caption[]{AP scores of different saliency detection methods on MSRC datasets. Overall, the proposed approach performs better in most situations.}
\label{AP}
\end{figure}

In addition, we make a quantitative performance comparison with some other recently proposed co-saliency approaches over the
MSRC dataset, including ESMG, ESMG-S (the variant of ESMG without group interactions), SACS, and CoDW.
Since these approaches have no open source codes, we have to directly quote their quantitative results (only having AP scores),
which are provided in the work~\cite{zhang2016detection}.
As shown in Figure~\ref{AP}, our approach achieves the second best performance in co-saliency detection, and
is also comparable to the best CoDW approach (involving many stages and refinement postprocessing operations
like manifold rankings).
In contrast, our approach is straightforward, end-to-end, and without any postprocessing. Thus,
it is a promising choice in practice.

\begin{table}[t] \scriptsize 
\vspace{0.3cm}
\renewcommand{\tabcolsep}{3pt}
\centering
    \resizebox{1\linewidth}{!}{
    \begin{tabular}{|c|c|c|c|c|c|c|}\hline
    Dataset &       & Ours  & Ours\_S  & CBCS  & CBCS\_S   & CSHS    \\
    \hline
\multirow{3}[0]{*}{iCoseg} & mF  & \textbf{0.6983}  & 0.6935  & 0.6885  & 0.6443  & 0.5288   \\
          & AUC   & 0.9497  & 0.9256  & 0.9294  & 0.9106  & \textbf{0.9530}     \\
          & MAE   & \textbf{0.1018}  & 0.1343  & 0.1922  & 0.1517  & 0.1102     \\
    \hline
\multirow{3}[0]{*}{MSRC} & mF  & \textbf{0.5952}  & 0.5671  & 0.5206  & 0.5057  & 0.4612  \\
          & AUC   & 0.6997  & 0.6799  & \textbf{0.7030}  & 0.6981  & 0.6813  \\
          & MAE   & \textbf{0.2238}  & 0.2534  & 0.3677  & 0.2912  & 0.2587   \\
    \hline
\multirow{3}[0]{*}{Cosal2015} & mF  & \textbf{0.6084}  & 0.5512  & 0.5130  & 0.4942  & 0.4898  \\
          & AUC   & \textbf{0.8954}  & 0.8744  & 0.8261  & 0.8251  & 0.8521  \\
          & MAE   & \textbf{0.1434}  & 0.1611  & 0.2268  & 0.1980  & 0.1883   \\
    \hline
    \end{tabular}
}
  \caption{Comparison of mean F-measure using adaptive threshold (mF), AUC scores and MAE scores (smaller better). Our approach achieves the best performance w.r.t. all these metrics in most cases.}
\label{tab:comp}%
\end{table}

\subsection{Analysis of Proposed Approach}

As illustrated in Figure~\ref{figsaliency}, our method obtains more robust and complete salient regions. The boundaries of the salient regions are more clear, and in most examples, the proposed approach properly filters the background information. The comparison between our single image model and the group-input group-output model demonstrates the effectiveness and important role of our group-wise feature representation as well as the collaborative learning strategy for the group-wise and single image features, when compared to the single model approach, the proposed one gives results where the common objects are enhanced and made brighter whereas the different objects are weakened and made dimmer. Meanwhile, we compute the average performance for group-wise model and single image model over all the datasets with respect to F-measure, MAE, and AUC. Overall, our group-wise model respectively achieves $0.6340$, $0.8477$ and $0.1563$ on F-measure, AUC, and MAE, and the single image model correspondingly achieves $0.6039$, $0.8266$, and $0.1829$. This effect is also most clear in Figure~\ref{figsaliency} on the apples image group where the trees that were detected as salient by the single image model were erased by the group-wise approach because it is not common to all the images of the group.

\section{Conclusion}

In this paper, we propose a unified deep co-saliency approach for co-salient detection made as a fully convolutional network with group input and group output. It takes a data-driven learning pipeline for capturing the collaboration and consistency intra image group, and subsequently builds an end-to-end learning scheme for explore the intrinsic correlations between the tasks of individual image saliency detection and intra group saliency detection. Through collaborative learning from the co-saliency image group, the deep co-saliency model obtained the capability of capturing the information of both the shared and unique characteristics of each image within the image group and effectively modeled the interaction relationship between them. The experimental results demonstrated that the proposed approach performs favorably in different evaluation metrics against the state-of-the-art methods.

\section*{Acknowledgments}
This work was supported in part by the National Natural Science Foundation of China under Grant U1509206 and Grant 61472353, in part by the Alibaba-Zhejiang University Joint Institute of Frontier Technologies.

\bibliographystyle{named}
\bibliography{ijcai17}

\end{document}